# Temporal Reasoning in AI Systems

**Abhishek Sharma**

3840 Far West Blvd, Austin, TX 78731
abhishek81@gmail.com

**Abstract**

Commonsense temporal reasoning at scale is a core problem for cognitive systems. The correct inference of the duration for which fluents hold is required by many tasks, including natural language understanding and planning. Many AI systems have limited deductive closure because they cannot extrapolate information correctly regarding existing fluents and events. In this study, we discuss the knowledge representation and reasoning schemes required for robust temporal projection in the Cyc Knowledge Base. We discuss how events can start and end risk periods for fluents. We then use discrete survival functions, which represent knowledge of the persistence of facts, to extrapolate a given fluent. The extrapolated intervals can be truncated by temporal constraints and other types of commonsense knowledge. Finally, we present the results of experiments to demonstrate that these methods obtain significant improvements in terms of Q/A performance.

## Introduction and Motivation

Modern cognitive systems are expected to reason about a complex and dynamic world. Commonsense temporal reasoning plays an important role in reasoning about a continuously changing environment. Since even the largest knowledge bases (KBs) lack complete knowledge of the outside world, cognitive systems must employ heuristic reasoning and estimate how long a given state is likely to persist. For example, if Fred falls down and sprains his ankle, he is expected to be in pain for a short while. If Fred's younger brother was born in 2013, it is to be expected that he will start to walk and talk within a few years. How can an AI system with incomplete information reason about these events? How long do facts persist? How do individuals change over time and what factors are relevant for understanding inter-individual differences? What events are expected to occur and when do they occur? What changes their probabilities of occurrence?

Many knowledge-based systems do not represent time correctly and they do not perform sophisticated temporal reasoning. In particular, the problem of temporal projection (i.e., the persistence of facts) has not received sufficient attention in the knowledge-based systems community. For this reason, many cognitive systems have limited deductive closure because existing facts (true at a given time point) cannot be extrapolated to answer questions at other time points. What types of knowledge representation and reasoning schemes are needed to solve these problems?

In this study, we use techniques from discrete time survival analysis to answer these questions. We discuss how events start and end risk periods during which an interval for a fluent (a time-varying property of the world) could be terminated. After identifying the starting points of the risk periods, we use hazard functions to construct an interval during which a time-dependent sentence is highly likely to be true. These intervals can be truncated by temporal constraints and other types of commonsense knowledge. We discuss how different temporal properties of predicates and collections can be used to infer hazard functions. Next, we explain how time (in)dependent covariates can be specified to scale the parameters of hazard functions. The role of event calculus in the context of temporal projection is then discussed, which is often used to maintain correct temporal intervals for fluents in databases. However, some of the predictions made by event calculus might be erroneous if relevant information about the expected duration for which fluents hold is not available. Thus, we discuss how survival functions can be used to alleviate this problem. Next, we explain how the parameters of these survival distributions can be learnt from data.

This paper is organized as follows. We start by discussing related work. Then we provide a brief introduction to the Cyc Knowledge Base (Cyc KB). Next, we discuss our temporal projection algorithm and the knowledge representation it requires. We conclude by discussing our experimental results and plans for future research.

## Related Work

In AI, the problem of temporal projection has been studied in the context of the frame problem [Hanks & McDermott 1986]. The use of survival functions for temporal projection in AI was initiated by [Dean & Kanazawa 1988] and it was extended by other researchers [Tawfik & Neufeld 2000]. Issues related to temporal abstraction in the medical domain were discussed in [Shahar 1997] and temporal-semantic properties were introduced in [Shoham 1987]. However, none of these addressed the problem in the context of improving Q/A performance of large-scale cognitive systems. The work presented in the present study is closest to that discussed by [Lenat 1998] and [Singer & Willet 2003]. Previous studies have developed probabilistic models for temporal and probabilistic reasoning [Hanks & McDermott 1993], but few researchers have focused on discussing knowledge-representation and reasoning issues that would help large logic-based AI systems to perform robust temporal projection and answer a wide range of answers.

## Background

In this section, we summarize the key conventions used by Cyc [Lenat & Guha 1990, Matuszek *et al.* 2006]. Cyc represents concepts as collections. Each collection is a kind or type of thing, the instances of which share a certain property, attribute, or feature. For example, *Cat* is a collection of all cats, and only cats. Collections are arranged hierarchically by the *genls* relation. *(genls <sub> <super>)* means that anything that is an instance of <sub> is also an instance of <super>. Predicates are also arranged in hierarchies. (*genlPreds* <s> <g>) means that <g> is a generalization of <s>. The collection *Situation* is an important part of Cyc's ontology. Each instance of *Situation* is a state or event that comprises one or more objects with certain properties or certain relations with each other. *Event* and *StaticSituation* are notable specializations of *Situation*. In Cyc, contexts are represented as *microtheories*. Each microtheory groups a set of assertions together that share some common assumptions. The time dimension of contexts can be specified [Lenat 1998]. For example, the following sentence represents the fact that Tony Greig was a cricketer between 1972 and 1977.

Monadic microtheory[1]: PeopleDataMt
Time dimension:
   (TimeIntervalInclusiveFn (YearFn 1972) (YearFn 1977))
Sentence: (isa TonyGreig-Cricketer Cricketer)

In Cyc, instances of *TimeDependentCollection* and *TimeDependentRelation* can be used to represent fluents. Each instance of *TimeDependentCollection* is a collection where the membership changes over time. For example, *Professor* is a time-dependent collection, whereas *Integer* is not. Similarly, *TimeDependentRelation* is the collection of all relations such that nothing remains in their eternal extensions. For example, *'owns'* and *'likesAsFriend'* are time-dependent relations. Inference and learning in Cyc-based systems can be guided by several heuristics [Sharma et al 2016, Sharma & Goolsbey 2017, Sharma & Goolsbey 2019, Sharma & Forbus 2010, Forbus et al 2009].

## Risk Periods and Discrete-Time Survival Analysis

Let us assume that $T_0$ was the last time point when fluent P was known to be true and no events are known to have occurred that affect the persistence of P[2]. If we lack perfect knowledge, we need to extrapolate and find a reasonable assessment of whether P would be true at time $T_0+\Delta$. Systems that use probabilistic logic might compute and use Prob (holds (P, $T_0+\Delta$)) directly. Other systems that use traditional forward and backward inference might compute Prob (holds (P, t)) for different values of t, and derive an interval ([T1, T2]) around $T_0$ in which the fluent P is highly likely to be true. If we assert that P is true in [T1, T2], this assertion can participate in forward and backward inference in the same manner as any other assertion in the KB[3].

Consider the following sentences:
  (isa Fred BiologicalLivingObject)             ... (A1)
  (isa Fred Married)                                 … (A2)
  (isa Fred MicrosoftEmployee)               … (A3)

Let us assume that A1, A2, and A3 were true in the year 1990, and we need to find the likelihood of their being true in the year 1992. Since the state represented by A1 is terminated by a death event, we have to find the probability of the occurrence of a death event in the time interval 1990–1992. Similarly, the likelihood of the truth of A2 in 1992 depends on Fred's age and on the duration for which he has been married. Therefore, the process of computing these likelihoods involves the following steps.

(a) **Identification of the starting points of *risk periods* when a state could be terminated**: Since states can be terminated as soon as they are started, this starting point is specified by the event that initiates the state. For example, the starting points of the risk periods for A1, A2, and A3 are given by the birth date, the wedding date, and the hiring date, respectively. This knowledge can be

---
[1] A monadic microtheory is a temporally or otherwise unqualified microtheory.

[2] The case when relevant events are known to have occurred is discussed in the next section.
[3] Obviously, temporal projection should not be used recursively. When contradictory information arrives, the time intervals of these assertions are updated (discussed below).

represented in the following format:
(eventTypeStartsRolePlayersRiskPeriodForState
 EVENT-TYPE ROLE STATE). This states that an event of type 'EVENT-TYPE' starts a risk period in which the individual who plays the role 'ROLE' in the event[4] could become an instance of 'STATE'. This information can be used to derive sentences such as A5.
(eventStartsRiskPeriodForSentence
     WeddingEvent-001 (isa Fred Divorced))      ….(A5)
If Fred's wedding event happened on July 1, 1988, then we can derive A6.
(startingPointOfRiskPeriodOfSentence
   (isa Fred Divorced) (DayFn 1 (MonthFn July (YearFn 1988)))) ….(A6)
Assertions such as A6 can be derived for each of Fred's weddings.

(b) **Specification of the *hazard function***: Given the starting point of the risk period, we can calculate the probability that a state-terminating event would occur in a given time interval. Let us divide continuous time into a sequence of time intervals $(0, t_1], (t_1, t_2], ...$, and so forth, and let $(t_{j-1}, t_j]$ be the $j^{th}$ time interval. Then, if T is the discrete random variable that indicates the time interval during which a state terminating event occurs, then the discrete hazard, $h_j$, is the conditional probability that a randomly selected individual will experience the state-terminating event in the $j^{th}$ time interval, given that he/she has not experienced it in preceding time intervals [Singer & Willett 1993, Singer & Willett 2003]:

$$h_j = \Pr(T = j \mid T \geq j) \quad ...(E1)$$

Thus, it follows that:
$$\Pr(T > k) = (1-h_1)(1-h_2)...(1-h_k)$$
$$= \prod (1-h_j) \quad ... (E2)$$
$$\Pr(T = k) = (1-h_1)(1-h_2)...(1-h_{k-1}).h_k$$
$$= h_k.\prod (1-h_j) \quad …(E3)$$

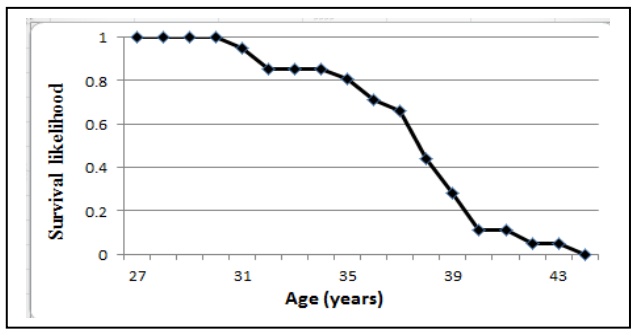

Figure 1: Likelihood of truth of sentences of type (isa <ins> Cricketer) as a function of age.

Given sentences like A6, we will find the last known starting point of risk period. Then expression E2 can be used to calculate the likelihood that sentences such as A1, A2, and A3 persist at the time $T_0+\Delta$. An example of such survival likelihoods (estimated by Cyc) is shown in Figure 1. We see that the professional career of cricketers starts ending when they are in their mid-30s, and virtually all of them retire before they are 45 years old. Each instance of the time-dependent collection and relations will potentially require different types of distributions and parameter values. Given the type of these distributions and the values of their associated parameter values, Cyc will use them automatically to extrapolate fluents. However, some distributions can be specified more easily if we use the following temporal semantic properties. (a) **Initial Collections**: In CycL, an instance, COL, of InitialCollection is defined such that any instance of COL starts as an instance of it, and if the instance changes so it is no longer an instance, it can never become an instance again. "NeverSchooled" and "FemaleInfant" are instances of InitialCollection. Thus, if (isa Tom NeverSchooled) was true on January 1, 1980, Cyc would use the property of the initial collection and assert that Tom was an instance of that collection from his birth until January 1, 1980. (b) **Terminal Collection**: Similarly, in CycL, an instance, COL, of TerminalNonInitialCollection, is a collection such that when a thing becomes an instance of COL, it must remain so while it exists. The collection "HumanAdult" and "Graduate" are examples of terminal collection. (c) **Bidirectional Projection**: In some cases, fluents persist as long as the individual exists[5] (e.g., FemaleHuman). Cyc's temporal projection module uses these properties.

How does this analysis change when we know other facts that affect the persistence of a given sentence? In the present study, we consider two ways in which this change might occur.

(a) **Temporal Constraints**: Consider the following sentences:
Microtheory: PeopleDataMt
Time Interval: (YearFn 1998)
Sentence: (isa JohnMcCarthy-ComputerScientist Professor)   …(A7)

Microtheory: PeopleDataMt
Time Interval: (YearFn 2001)
Sentence: (isa JohnMcCarthy-ComputerScientist RetiredPerson) ...(A8)
Given a sentence such as A7, Cyc's temporal projection module would try to extrapolate and construct a time interval around 1998 during which John McCarthy was a professor. However, a sentence such as A8 suggests that he

---
[4] For the sentence A2, we want to state that the wedding event starts a risk period in which the individual who plays the role "groom" could enter the state "Divorced". A similar sentence would be needed for the bride.

[5] For simplicity, we ignore the possible occurrence of relatively rare gender change events. Temporal projection of sentences of type (isa <ins> COL) where COL is a specialization of BiologicalLivingObject, often involves reasoning about the life expectancy of an individual. This is done with the help of hazard functions such as E1, which are calculated using mortality rates.

was not a professor in 2001. We represent this information using the following sentence.

(followingStageTypes Professor RetiredPerson)…(A9)

Sentence A9 represents that a person's life as a professor ends before his life as a retired person begins[6]. Let P represent the fluent "John McCarthy is a professor." Then, based on only A8 and A9, Cyc can infer that John McCarthy was a professor at time point T if the following conditions are met: (a) Prob (P, T) > threshold, and (b) ¬(Prob (Q, T) > threshold), where Q is an assertion that is incompatible[7] with P.

(b) **Covariates**: Temporal constraints can help us to represent the incompatibility of two assertions that mainly limit the time interval during which assertions are deemed to hold. However, additional information about an individual might change the rate of decay of persistence. For example, a sentence such as A10 is a time-dependent covariate for A11.
Microtheory: PeopleDataMt
Time interval: 1970 to 1975.
Sentence: (isa Fred (FrequentPerformerFn Smoking))   ...(A10)
Sentence: (isa Fred BiologicalLivingObject)   …(A11)

Other covariates (e.g., gender) can be time independent. To scale the hazard when such covariates are present, an expression such as E4 is used.

$$h(i) \leftarrow 1/ (1+e^{-\alpha(i)} e^{(-\Sigma \beta(i) * X(i))}) \quad …(E4)$$

Recall that $h(i)$ is the hazard for the $i^{th}$ time interval. Here, each $X(i)$ is a Boolean variable that corresponds to a covariate such as A10. The value of $X(i)$ is 1 if the covariate is present in the $i^{th}$ time interval, but zero otherwise. The parameters $\beta(i)$ represent the strengths of the covariates. When a sentence such as A11 is true in an interval $i$, the value of $h(i)$ increases, which leads to a sharper decline in the probability of the persistence of A11. When all of the values of $X(i)$ are zero, the hazard function for the $i^{th}$ time interval reduces to E5.

$$h(i) \leftarrow 1/ (1+ e^{-\alpha(i)}) \quad …(E5)$$

Expression E5 represents the baseline hazard function, which is determined by the parameter $\alpha(i)$. Information about covariates can be represented by assertions such as the following.
(timeDependentCovariateForCollection   BiologicalLivingObject (FrequentPerformerFn Smoking) 0.3)   …(A12)

---

[6] In CycL, the collection RetiredPerson is the collection of people who have retired permanently.
[7] In this case, sentences such as "John McCarthy is a retired person." and "John McCarthy is an infant." are incompatible with P. Such incompatibility can often be inferred by using 'disjointWith' assertions. For instance, (disjointWith HumanInfant Professor) holds.

The last argument in A12 is used to specify the value of $\beta(i)$. Similar assertions are used for time-independent covariates.

Hazard functions can also be used to infer the persistence of time-dependent predicates. However, in some cases, we need an additional vocabulary for plausible inferences. Consider the following sentences:
(owns AlbertEinstein Car-780)   …(A13)
(owns AlbertEinstein Toothbrush-392)   …(A14)
The sentences A13 and A14 should have different decay rates. Therefore, we need to specify hazard functions for handling sentences where an instance of a certain collection appears in a given argument position in a predicate[8]. If the known specific conditions are not satisfied for a given assertion, then we must employ generic hazard functions that apply to a given predicate.

## Event Calculus in Temporal Projection

The inferences made by the methods discussed above are non-monotonic in nature. When contradictory information arrives, we can use methods based on both event calculus and survival analysis to maintain a consistent KB. Let us start with the basics of event calculus. Briefly, fluents are considered true at a time point if they have been initiated by an event previously, but have not been terminated in the meantime. Similarly, a fluent is false if it has been terminated previously but has not been initiated in the meantime [Miller & Shanahan 2002]. In Cyc, domain-dependent axioms are written to derive sentences such as the following, which represent knowledge about situations that start and end time intervals for fluents.
(situationStartsIntervalForSentence  s f)   …(A15)
(situationEndsIntervalForSentence t g)   …(A16)
The first sentence above denotes that situation *s* starts a time interval during which the fluent *f* holds. The core persistence axiom employed in simplified event calculus is as follows [Sadri & Kowalski 1995].

Holds(P, T) ← Happens (E1, T1) and Initiates (E1, P) and
   T1 < T and ~∃ E2, T2 (Happens (E2, T2) and
   Terminates (E2, P) and T1 < T2 < T)   … (A17)

However, the original event calculus description [Kowalski & Sergot 1986] suggested that extra application-specific rules can be added to infer whether an interval for a fluent has been broken if the known events are too far apart for it to hold continuously. This problem can be solved if we combine the event calculus with the survival analysis-based methods discussed above. In step 1a of the algorithm shown in Figure 2, we use A17 to construct an interval [T1, T2] for a fluent P. However, the intervals created in step 1b are preferred[9] if they are subsumed by [T1, T2]. Since hazard functions represent knowledge about how long states persist, this helps us to avoid the error caused when excessively wide time intervals are produced due to incomplete knowledge of the outside world. Since many events that start/end intervals for P might be known to us, we create a timeline of events and choose E1 and E2 such that following condition is satisfied.

---

[8] For A13, the relevant hazard function represents our knowledge of how long automobiles are owned.

happens (E1, T1), happens (E2, T2),  T1 < T < T2,
Initiates(E1, P), Terminates (E2, P),
$\sim \exists$ E3 such that (happens(E3, T3), Initiates (E3, P),
T1 < T3 < T < T2) , $\sim \exists$ E4 such that (happens(E4, T4),
Terminates (E4, P),    T1  < T < T4< T2)            …(C1)

In step 2, when all known events start time intervals for P, we use C2 to choose an event E1, and use the hazard function to create an interval that extends forward from E1.

happens(E1, T1), Initiates (E1, P) and $\sim \exists$ E2 such that (happens (E2, T2) and T1 < T2 < T and Initiates (E2, P))
…(C2)
happens(E1, T1), Terminates (E1, P) and $\sim \exists$ E2 such that (happens (E2, T2) and T < T2 < T1 and Terminates (E2, P)).
…(C3)

Conditions C1, C2 and C3 help us identify events that are most temporally proximate to P. When all known events end intervals for the given fluent, we use C3 to choose the closest event and create an interval that extends backwards from it (step 3). If the intervals created by these events do not subsume T, then the truth of P at T is unlikely to be related to their occurrence and step 4 is executed. In step 4, we use the hazard functions and create a time interval [T5, T6]. In step 5, we look at different constraints that might be applicable and we truncate the interval if necessary.

**Non-survival Analysis-related Vocabulary**: Although survival analysis provides a natural framework for temporal projection, some fluents are less amenable to monotonically decreasing persistence likelihoods. For example, consider the following sentence:
(isa BillClinton UnitedStatesPresident)            …(A17)
It is plausible to infer that a sentence such as A17 would be true for four years from the inauguration date of the president. To handle these cases, we can derive sentences such as A17.
(statePersistsForDurationFromDate
 (isa BillClinton UnitedStatesPresident)
 (DayFn 21 (MonthFn January (YearFn 1993))) (YearsDuration 4)) (A17)

These assertions are derived from domain-specific axioms (e.g., axioms about presidential inauguration) and they are treated as default statements. They can be overridden when contradictory information[10] is available and assertions such as A16 are derived.

---
**Algorithm**: TemporallyProject
**Input**: A Knowledge Base, KB
   A fluent, P, true during a time interval, T.
   A likelihood threshold, α.
**Output**: A time interval [T1, T2] that subsumes T, such that if T3 ε [T1, T2], then Prob (P, T3) > α.
1. If events E1 and E2 exist such that they start and end intervals for P respectively and satisfy condition C1, then execute 1a–1c, else goto step 2.
   (1a) Use A17 to construct an interval [T1, T2] that subsumes T for P.
   (1b) Use the hazard function for P to construct another interval [T3, T4] for P.
   (1c) If [T1, T2] subsumes [T3, T4], then return [T3, T4], else return [T1, T2].
2. If all known events start intervals for P, use condition C2 to choose an event, and use the relevant hazard function to create an interval [T5, T6] that extends forward from the event. If T is subsumed by [T5, T6] then return [T5, T6] else goto step 4
3. If all known events terminate intervals for P, use condition C3 to choose an event and use the relevant hazard function to create an interval [T5, T6] that extends backward from the event. If T is subsumed by [T5, T6] then return [T5, T6] else goto step 4.
4. Use hazard functions to create a time interval [T5, T6] that subsumes T.
5. If applicable, use temporal constraints to truncate [T5, T6]. Return the resulting time interval.

**Figure 2: Algorithm used for temporal projection.**

---

**Learning:** How can we learn the parameters of the probability distributions discussed above? We recall that the definitions of hazard functions are reliant on identifying the time interval during which the state terminating event occurs. Therefore, the task of learning the hazard functions requires information in the following format [Singer & Willet 2003].
(i) **A time period variable, $j$**: We need to divide the time period since the start of the risk period into intervals with a suitable length. The time period variable specifies the time period $j$ of the record. For example, let us consider sentence A11. Fred's birth event starts the risk period and intervals with a length of one year are suitable for estimating the likelihood of its persistence. Therefore, the fourth year of Fred's life would correspond to the fourth time interval in our dataset. (ii) **Values of covariates in each time interval**: For each time interval, $j$, we need to obtain the value of the covariate (i.e., the truth of sentences such as A10) in that time interval for the given individual. (iii) **Value of EVENT($i, j$)**: The variable EVENT($i, j$) indicates whether the state terminating event for individual $i$ occurred during the time period $j$. Given a set of such records, the likelihood of observing the data is given as follows.

---
[9] In Figure 2, the intervals created by hazard functions are such that if T lies in the interval then Prob(holds(P, T)) > α, where α is an input to the algorithm.

[10] For fluents involving US presidents, assassination, impeachment or resignation events will terminate intervals for sentences such as A17.

$$L = \prod_i \prod_j h(t_{ij})^{EVENT(i,j)} (1-h(t_{ij}))^{(1-EVENT(i,j))}$$

The values of the hazard functions can be calculated by minimizing L.

| Query set | Mode | #Queries | % Correct | Improvement w.r.t. Mode 1 |
|---|---|---|---|---|
| 1 | M1 | 100 | 28% | - |
|   | M2 | 100 | 56% | 100% |
| 2 | M1 | 3661 | 39% | - |
|   | M2 | 3661 | 57% | 46% |
| 3 | M1 | 423 | 29% | - |
|   | M2 | 423 | 70% | 141% |
| 4 | M1 | 2616 | 43% | - |
|   | M2 | 2616 | 63% | 47% |
| 5 | M1 | 533 | 26% |   |
|   | M2 | 533 | 34% | 31% |
| Total | M1 | 7333 | 39% |   |
|   | M2 | 7333 | 58% | 49% |

Table 1: Experimental Results

## Experimental Evaluation

To assess the validity of these concepts, we conducted a set of experiments. Five query sets were selected based on the availability of temporally qualified ground facts and their relevance to temporal projection. Every query was in the form: "Sentence s was (will be) true at time T"[11], where s was a fully bound fluent of the type '(isa <ins> <col>)'. For each of these query sets, we measured the Q/A performance in two modes, as follows. **Mode M1**: In this mode, we aimed to simulate the performance of a traditional reasoning system with no temporal projection module. In addition to simply looking up the KB and using the predicate generalization hierarchy, the inference engine performed a temporal subsumption[12] test during reasoning. **Mode M2**: In the second mode, in addition to the reasoning done in mode M1, we enabled temporal projection methods discussed in this paper. The results of these experiments are shown in Table 1. We see that there has been significant improvement in Q/A performance in all query sets[13]. The overall improvement in performance with respect to M1 is 49%. The results are statistically significant ($p < 0.05$).

## Conclusions and Discussion

Commonsense temporal reasoning is a core problem for cognitive systems. To improve commonsense reasoning in general and Q/A performance in particular, modern AI systems must find better ways to reason about how long facts persist. In this study, we proposed different types of knowledge representation schemes, which may help such systems to perform robust temporal projection. The specification of risk periods and hazard functions facilitates the calculation of plausible intervals for fluents. Temporal constraints help to limit intervals and the presence of covariates can scale the parameters of the hazard functions. Event calculus can be integrated with survival analysis to alleviate some of the problems caused by incomplete knowledge. Experiments on a set of over 7000 queries shows that the methods proposed here lead to 49% improvement in Q/A performance on average.

Our results suggest the following future areas of research. First, we have found that in the case of many time dependent collections (e.g., Entertainer, Singer), once people enter these states; the fluent persists until the end of their active life. Therefore, we would like to reason about when individuals are likely to enter a given state, and how they might make transitions among states. Previous research into multi-state processes might be relevant in this context [Crowder 2012]. Second, we would like to extend this research to ensure that we can reason about recurrent and periodic events (e.g., sleeping or going to a grocery store). Finally, the capability to reason about probabilistic effects of events, and estimating the likelihood of event occurrence in a given time interval would be very useful for further improving the results shown in Table 1.

---

[11] We included a query only when it could not be answered easily by considering the lifespan of individuals. For instance, a query like "Was John McCarthy a professor in the year 1850?" would be excluded from our experiments. On the other hand, a query like "Was Marvin Minsky a professor in the year 1984?" would be included in our query sets.

[12] A sentence s is considered true during interval T1 if it is known to be true during interval T2, and T2 subsumes T1.

[13] The number of queries in query sets is different due to non-uniform distribution of facts in KB. For example, Cyc KB knows much more about movie actors than about cricketers.